\documentclass{article}
\usepackage[numbers]{natbib}

\usepackage[preprint]{nips}

\usepackage[utf8]{inputenc} \usepackage[T1]{fontenc}    \usepackage{hyperref}       \usepackage{url}            \usepackage{booktabs}       \usepackage{amsfonts}       \usepackage{nicefrac}       \usepackage{color}
\usepackage{epsfig}
\usepackage{amsthm}
\usepackage{amsmath}
\usepackage{float}
\usepackage{graphicx}
\usepackage{wrapfig}
\usepackage{makecell}
\usepackage{caption}
\usepackage{subcaption}
\usepackage{bm}
\usepackage{multirow}
\usepackage{microtype}
\usepackage{mathtools}
\usepackage{algorithm}
\usepackage{algorithmic}
\usepackage{listings}
\usepackage{enumitem}
\usepackage{xspace}

\newcommand{\nop}[1]{}
 \newcommand{\modelname}{FIT\xspace}
\newcommand{\modelnamear}{FIT-AR\xspace}

\title{FIT: Far-reaching Interleaved Transformers}

\author{
   Ting Chen$^\dagger$\thanks{$^\dagger$Correspondence to: \texttt{iamtingchen@google.com}} \\
   Google Deepmind \\
   \And
   Lala Li \\
   Google Deepmind \\
}
\renewcommand\footnotemark{}

\begin{document}

\maketitle

\begin{abstract}

We present \modelname: a transformer-based architecture with efficient self-attention and adaptive computation.
Unlike original transformers, which operate on a single sequence of data tokens, we divide the data tokens into groups, with each group being a shorter sequence of tokens.
We employ two types of transformer layers: local layers operate on data tokens within each group, while global layers operate on a smaller set of introduced latent tokens. These layers, comprising the same set of self-attention and feed-forward layers as standard transformers, are interleaved, and cross-attention is used to facilitate information exchange between data and latent tokens within the same group.
The attention complexity is $O(n^2)$ locally within each group of size $n$, but can reach $O(L^{\frac{4}{3}})$ globally for sequence length of $L$. The efficiency can be further enhanced by relying more on global layers that perform adaptive computation using a smaller set of latent tokens.
\modelname is a versatile architecture and can function as an encoder, diffusion decoder, or autoregressive decoder. We provide initial evidence demonstrating its effectiveness in high-resolution image understanding and generation tasks. 
Notably, FIT exhibits potential in performing end-to-end training on gigabit-scale data, such as 6400$\times$6400 images, or 160K tokens (after patch tokenization), within a memory capacity of 16GB, without requiring specific optimizations or model parallelism.
\footnote{Code at https://github.com/google-research/pix2seq}

\end{abstract}

 \section{Introduction}

Transformer~\cite{vaswani2017attention} is a neural network architecture designed to operate on a set of data tokens, such as text tokens~\cite{radford2019language} or image patch tokens~\cite{dosovitskiy2020image}. It employs self-attention mechanism that enables \textit{all-to-all} information exchange among the tokens, resulting in an $O(L^2)$ complexity.
Although transformers have demonstrated success in various domains, their quadratic complexity poses limitations when dealing with longer sequences. Efforts have been made to address this challenge, but the full quadratic attention mechanism remains the most effective and commonly used, particularly for shorter sequences.

To leverage quadratic attention for long sequences, we draw inspiration from how natural data can be organized into groups. For instance, text tokens in a book can be grouped into chapters, while patch tokens in an image can be organized into blocks or windows. Within each group, we can employ a high-bandwidth communication channel utilizing quadratic attention, while across groups, a lower-bandwidth channel with meaningful compression may suffice. This approach of dividing data tokens into groups or segments has been successfully applied in several existing works~\cite{zaheer2020big,vaswani2021scaling,han2021transformer,liu2021swin,li2022exploring}. However, the mechanism for coordinating local (intra-group) and global (inter-group) information processing remains under-explored.

In this work, we carefully design a mechanism that coordinates local and global processing efficiently. This is achieved by first 
introducing a small set of latent tokens~\cite{jaegle2021perceiver,jabri2022scalable} for each group. We further interleave two type of transformer layers, one for processing data tokens with local/window attention and the other for processing latent tokens with global attention.
Cross-attention is used to route information between the data tokens and latent tokens within the same group.
A single forward pass of the network involves iterative updates of both data tokens and latent tokens, ensuring local and global information to be sufficiently integrated.

We evaluate the proposed architecture on high-resolution image understanding (as image encoder) and generation (as diffusion decoder and autoregressive decoder), and provide initial evidences that the proposed architecture can serves as an efficient and effective extension of transformers for processing and generating long sequences.
 \vspace{-0.5em}
\section{Background and related work}
\vspace{-0.5em}

The primary sources of computational complexity in transformers arise from self-attention, which has a complexity of $O(L^2d)$, and the feed-forward network (FFN) with a complexity of $O(Ld^2)$. It is important to be aware of the typical scale of $L$ (sequence length) and $d$ (embedding dimension) that we may encounter. With the recent surge of large models, $d$ can vary from 1024 to 18432~\cite{brown2020language,chowdhery2022palm}. Longer contexts are often desired, leading to $L$ ranging from a few hundred to millions or even higher.
In cases where $d$ and $L$ are of comparable magnitudes, such as $d=4096$ and $L=2048$, approximations applied to the original quadratic attention may yield negligible gains as they usually shift the computational burden from $O(L^2d)$ to $O(Ld^2)$. Furthermore, optimized implementations of self-attention, such as FlashAttention~\cite{dao2022flashattention}, can improve efficiency and eliminate the need for $O(L^2)$ memory. Therefore, the full quadratic attention remains the most effective and efficient operation when dealing with relatively short sequences (e.g., in the order of hundreds, or thousands).

Considering the efficacy of full attention for shorter sequences and our objective of developing a single architecture capable of handling sequences of varying lengths, we will focus on reviewing techniques that integrate global attention into architectures with local quadratic attentions (such as window attention).
One simple approach is to leverage global attention for only a few transformer layers. Although this primarily reduces the constant factor in terms of computational complexity, it has demonstrated practical effectiveness in certain applications~\cite{brown2020language,li2022exploring}.
Another approach is to employ techniques such as shifted window~\cite{liu2021swin} or convolution~\cite{zhang2022nested} to propagate information across groups, with a disadvantage that it only affects adjacent groups at a time.
A more general approach involves the use of sparse/axial attention~\cite{child2019generating,ho2019axial}, mixed or learned attention patterns~\cite{zaheer2020big,zhao2021improved,kitaev2020reformer,roy2021efficient,ren2021combiner}. These methods allow for more flexible attention patterns, but may involve sparse operations that are not always accelerator-friendly.
Another family of methods~\cite{dai2019transformer,rae2019compressive,bulatov2022recurrent,hutchins2022block} incorporates recurrent mechanisms to connect transformers on local windows. However, similar to RNNs, these methods often require truncation due to reduced parallelism, which limits the sequence length during training.
Lastly, a recent approach~\cite{hua2022transformer} incorporates linear attention~\cite{katharopoulos2020transformers,wang2020linformer,choromanski2020rethinking} to bridge the information gap between local windows, but linear attention over all tokens in long sequences can still be expensive.
For a comprehensive overview of various efficient transformer variants, we refer readers to~\cite{tay2022efficient}.

Indeed, the aforementioned techniques are able to enhance long-range dependencies in local attention models. However, the communication channel across groups may be rigid, inefficient with accelerators, or not easily scalable. In contrast, we take a different approach by introducing a small set of adaptive latent tokens specifically designed for global attention. This allows for more flexible and efficient information exchange across groups. Additionally, we propose an interleaved mechanism, utilizing two types of transformer layers to encapsulate local and global processing, creating a dynamic interplay resembling top-down and bottom-up interactions~\cite{hinton2022represent}, and ensuring the model remains expressive and scalable.

An alternative approach to improving the computational efficiency of transformers is to shift the computation to a reduced set of tokens (i.e. more computation on shorter sequences). Traditional methods often employ fixed-pattern downsampling techniques such as max pooling or average pooling~\cite{dai2020funnel}, which have shown reasonable effectiveness. More recent approaches~\cite{lee2019set,guo2019star,gupta2020gmat,beltagy2020longformer,ainslie2020etc,jaegle2021perceiver,jaegle2021perceiverio,jabri2022scalable,hawthorne2022general,carreira2022hierarchical,han2021transformer,ryoo2021tokenlearner,gao2023sparseformer} explore the use of latent tokens that dynamically attend to data tokens and perform additional computations. Latent tokens are not tied to specific data tokens and remain small in number, thereby offering a compression effect and effectively handling redundancy or non-uniform information distribution in the data~\cite{jabri2022scalable}.
Our work builds upon similar ideas, but with a unique design. We incorporate grouping and local/window attention, seamlessly combining them to optimize efficiency for both encoding and generation tasks. 

More discussions on closely related architectures are provided in Appendix~\ref{app:related} for clarity. \section{Method}

\begin{figure}[t]
    \centering
    \includegraphics[width=0.9\textwidth]{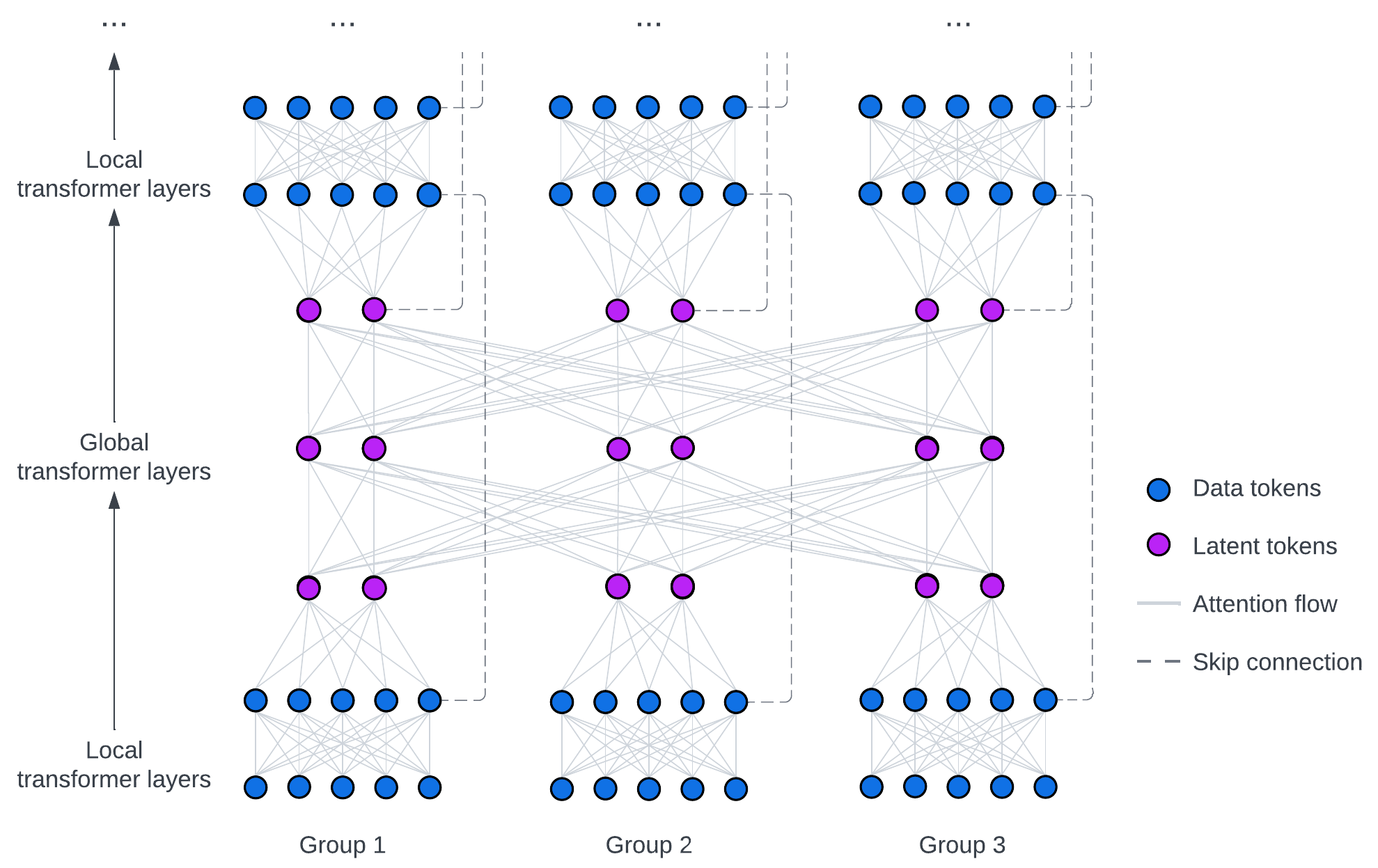}
    \caption{Illustration of the basic \modelname architecture. During the forward pass, the local transformer layers operate on the data tokens within each group independently and concurrently. Subsequently, the latent tokens selectively attend to the data tokens through cross attention. The latent tokens then undergo processing by the global transformer layers. Following this, the data tokens retrieve contextualized information from the latent tokens via cross attention.
    This process represents one block of forward processing. Multiple blocks are interleaved, alternating between the local and global transformers, to ensure a comprehensive mixture of information across the model.
}
    \label{fig:fit}
\end{figure}

We give a quick overview of key concepts in the proposed architecture, termed \modelname, or FitTransformer.

\textbf{Groups.} Transformers operate on a set of data tokens, where the ordering of tokens is managed through positional encoding rather than their specific layout in computer memory. The input to transformers, denoted as $\bm x\in \mathbb{R}^{b\times L\times c}$, represents the input shape as (batch size, number of tokens, token dimension). To facilitate processing, we divide a single group of data tokens into multiple groups. This essentially reorganizes the input $\bm x$ into $\mathbb{R}^{b\times t\times n\times c}$, where the new shape represents (batch size, number of groups, number of tokens per group, token dimension) and $L=t\times n$. The process of data grouping is flexible and can be achieved by directly splitting or reshaping a sequence into sub-sequences. In the case of images, it involves blocking the image into sub-images, with each sub-image treated as a separate group.

\textbf{Data (local) tokens vs. latent (global) tokens:} In the context of \modelname, we distinguish between data tokens and latent tokens. Data tokens correspond to those used in standard transformers and are typically associated with specific data elements. For example, in the case of an image, a data token can represent a patch embedding vector~\cite{dosovitskiy2020image}. Even after undergoing transformations through the transformer layers, data tokens maintain their association with specific parts of the data.
On the other hand, latent tokens are a small set of additional tokens introduced, often represented as positional embeddings that are not directly tied to the underlying data at the beginning~\cite{jaegle2021perceiver,jabri2022scalable}. However, during the forward pass, latent tokens dynamically aggregate information and become associated with specific parts of the data. This adaptive process varies from example to example, allowing the model to form longer-term memory and effectively compress the information in the data tokens.

\textbf{Local transformer layers vs. global transformer layers:} Both local and global transformer layers share a similar structure, comprising a standard self-attention module followed by a feed-forward network. However, they operate on different sets of tokens within the model.
Local transformer layers are applied to data tokens within each group. These layers process the data tokens within their respective groups, allowing for localized information processing and capturing fine-grained relationships among the tokens within the group. It is worth noting that local transformer layers can be customized by replacing them with other architectural building blocks, such as convolutions, or simplifying them by removing the self-attention~\cite{jabri2022scalable}.
On the other hand, global transformer layers are responsible for processing the latent tokens across all groups. These layers enable the model to capture global dependencies and long-range relationships between different parts of the input.

\begin{figure}[t]
  \centering
\begin{minipage}[t]{0.91\textwidth}
\begin{algorithm}[H]
\small
\caption{\modelname architecture. More details are in Algorithm~\ref{alg:utils}.}
\label{alg:fit}
\definecolor{codeblue}{rgb}{0.25,0.5,0.5}
\definecolor{orange}{rgb}{0.96, 0.81, 0.375}
\definecolor{codekw}{rgb}{0.85, 0.18, 0.50}
\lstset{
  backgroundcolor=\color{white},
  basicstyle=\fontsize{8pt}{8pt}\ttfamily\selectfont,
  columns=fullflexible,
  breaklines=true,
  captionpos=b,
  commentstyle=\fontsize{8pt}{8pt}\color{codeblue},
  keywordstyle=\fontsize{8pt}{8pt}\color{codekw},
  escapechar={|}, 
}
\begin{lstlisting}[language=python]
def fit_transformer(x):
  """Computation defined on grouped data tensor of shape (b, t, n, c)."""
  b, t, n, c = x.shape  |~~~|# batch size, num of groups, num of tokens per group, dim.
  x += positional_encoding(x)  |~~~~~~~~~~~~~~~~~~~~~~~~~~~~~~~~~~~~~~~|# (b, t, n, c).
  latents = initialize_latents()  |~~~~~~~~~~~~~~~~~~~~~~~~~~~~~~~~~~~~|# (b, t, m, d).
  x = rearrange(x, `b t n c -> (b t) n c`)  |~~~~~~|# reshape to prepare for attention.
  latents = rearrange(latents, `b t m d -> (b t) m d`)
  
  for i in range(num_net_blocks):
    x = |\color{blue}local\_transformers|[i](x)  |~~~~~~~~|# layers applied in parallel for each group.
    latents = |\color{blue}l2x\_cross\_attn|[i](latents, x)
    latents = rearrange(latents, `(b t) m d -> b (t m) d`)
    latents = |\color{blue}global\_transformers|[i](latents)  |~~~~|# layers applied across all groups.
    latents = rearrange(latents, `b (t m) d -> (b t) m d`)
    x = |\color{blue}x2l\_cross\_attn|[i](x, latents)
 
  return x, latents
\end{lstlisting}
\end{algorithm}
\end{minipage}
\end{figure}

\subsection{The basic \modelname architecture}

Figure~\ref{fig:fit} illustrates the basic \modelname architecture that operate on data tokens of $\mathbb{R}^{b\times t\times n\times c}$ and latent tokens of $\mathbb{R}^{b\times t\times m\times d}$ where $m\ll n$. And Algorithm~\ref{alg:fit_ar} provides a pseudo-code for its implementation.
To better understand how the \modelname architecture connects to the standard transformer and other existing architectures, we examine several special cases of settings below.
\begin{itemize}[topsep=0pt, partopsep=0pt, leftmargin=13pt, parsep=0pt, itemsep=4pt]
\item If we set a single group for data tokens (i.e., no grouping), \modelname reduces to an architecture that closely resembles RIN~\cite{jabri2022scalable}. However, RIN does not use full attention among data tokens due to its computational cost for a large number of tokens. If we specialize it further, by only using a single block of local$\rightarrow$global$\rightarrow$local layers, it also resembles Perceiver IO~\cite{jaegle2021perceiverio}.\item If we set the number of groups equal to the number of data tokens (i.e., treating each data token as a separate group), \modelname becomes similar to the standard transformer, albeit with an additional per-token network that may not be necessary in this case.
\item Viewing the local transformer layers operating on data tokens within each group as a standard transformer, \modelname can be considered an augmentation of the standard transformer. It introduces extra global transformer layers that connect local segments and provide contextualized feedback, enhancing the model's expressive power.
\item Regarding the global transformer layers operating on latent tokens as a standard transformer, \modelname can be seen as an augmentation of the standard transformer through the introduction of a learned adaptive tokenization. This tokenization summarizes data tokens, which can be an already compressed patch embedding, into latent tokens, enabling more efficient and compact processing.
\end{itemize}

\subsection{Extending the basic \modelname for autoregressive modeling}

\begin{figure}[t]
    \centering
    \includegraphics[width=0.9\textwidth]{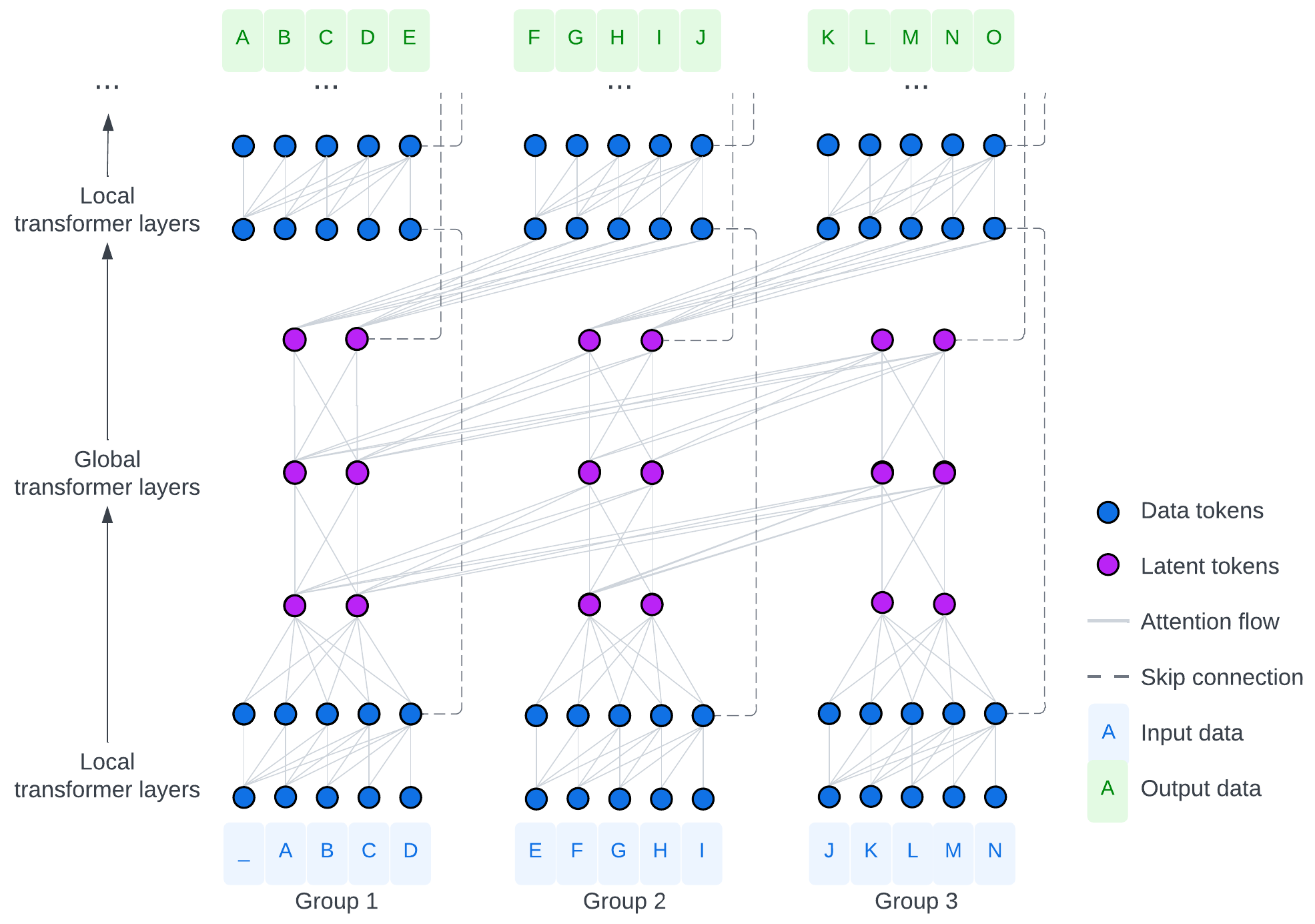}
    \caption{\label{fig:fit_ar}Illustration of the \modelnamear architecture for autoregressive modeling. 
    In contrast to the basic \modelname, this variant incorporates causal masks and shifting in cross-attention between data and latent tokens to prevent information leakage from future tokens to the past. }
\end{figure}

\begin{figure}[ht!]
  \centering
\begin{minipage}[t]{0.9\textwidth}
\begin{algorithm}[H]
\small
\caption{\modelname-AR architecture (training time). More details are in Algorithm~\ref{alg:utils}.}
\label{alg:fit_ar}
\definecolor{codeblue}{rgb}{0.25,0.5,0.5}
\definecolor{orange}{rgb}{0.96, 0.81, 0.375}
\definecolor{codekw}{rgb}{0.85, 0.18, 0.50}
\lstset{
  backgroundcolor=\color{white},
  basicstyle=\fontsize{8pt}{8pt}\ttfamily\selectfont,
  columns=fullflexible,
  breaklines=true,
  captionpos=b,
  commentstyle=\fontsize{8pt}{8pt}\color{codeblue},
  keywordstyle=\fontsize{8pt}{8pt}\color{codekw},
  escapechar={|}, 
}
\begin{lstlisting}[language=python]
def fitar_transformer(x):
  """Causal computation defined on data tensor of shape (b, t, n, c)."""
  b, t, n, c = x.shape  |~~~|# batch size, num of groups, num of tokens per group, dim.
  x += positional_encoding(x) |~~~~~~~~~~~~~~~~~~~~~~~~~~~~~~~~~~~~~~~|# (b, t, n, c).
  latents = initialize_latents() |~~~~~~~~~~~~~~~~~~~~~~~~~~~~~~~~~~~~|# (b, t, m, d).
  x = rearrange(x, `b t n c -> (b t) n c`)  |~~~~~~|# reshape to prepare for attention.
  
  for i in range(num_net_blocks):
    x = |\color{blue}local\_transformers|[i](x, causal_mask)
    latents = rearrange(latents, `b t m d -> (b t) m d`)
    latents = |\color{blue}l2x\_cross\_attn|[i](latents, x)
    latents = rearrange(latents, `(b t) m d -> b (t m) d`)
    latents = |\color{blue}global\_transformers|[i](latents, group_causal_mask)
    latents = rearrange(latents, `b (t m) d -> b t m d`)
    latents, latents_last = |\color{orange}shift\_latents|(latents) |~~~~~~~~~~~~~~|# ensure causality.
    x = |\color{blue}x2l\_cross\_attn|[i](x, latents)  |~~~~~~~~~~~~~~~~~~~~~|# latents in (b*t, m, d).
    latents = |\color{orange}shift\_back\_latents|(latents, latents_last)
  
  x = rearrange(x, `(b t) n c -> b t n c`)
  return dense(x) |~~~~~~~~~~~~~~~~~~~~~~~~~~~~~~| # logits for predicting next token.
\end{lstlisting}
\end{algorithm}
\end{minipage}\end{figure}

In autoregressive (language) modeling, it is crucial to prevent the flow of information from future data tokens into the past. This requirement can be easily achieved at the global transformer layers by adopting a block-wise causal mask, which allows full visibility among latents within a group but imposes a causal mask across groups.
However, in the basic \modelname architecture, there is a potential issue where information can unintentionally leak from future tokens into the past within the same group. To overcome this challenge, we introduce the concept of shifted latents between pushing and pulling information. Specifically, when data tokens in the $i$-th group push information to latent tokens of the same group, they are required to pull information from the latent tokens of the $(i-1)$-th group. This shifting mechanism ensures that information flows in a consistent and causal manner, preventing any inadvertent leakage of future information into the past.
This is illustrated in Figure~\ref{fig:fit_ar}, and pseudo-code for training \modelnamear architecture is given in Algorithm~\ref{alg:fit_ar}. In terms of inference, the model still decodes one token at a time autoregressively, but the presence of latent tokens summarizing preceding data tokens in \modelname can significantly improve decoding speed for long sequences, while also reducing memory usage.

\subsection{Complexity and Efficiency Analysis}

\modelname offers two notable efficiency improvements compared to standard transformers:
Firstly, with interleaved local and global attention, it significantly reduces the complexity of attention layers, going from a quadratic complexity of $O(L^2)$ to an optimal complexity of $O(L^\frac{4}{3})$.
Secondly, the architecture enables adaptive computation. By offloading the processing of local transformer layers to the global transformer layers, which operate on a smaller set of adaptive latent tokens, the overall computational cost is further reduced. These efficiency improvements in the proposed architecture make it well-suited for handling long sequences while maintaining computational tractability.
Table~\ref{tab:complexity} breaks down computation cost for both standard transformers and \modelname. And a detailed computational complexity analysis on attention operations can be found in appendix ~\ref{app:attn_complexity}.

\begin{table}[h]
    \small
    \centering
    \caption{Computation and complexity breakdown.
The basic computation units in both Transformer and \modelname are nearly identical, consisting of (dense) attention layers and feed-forward networks (FFN). Note that total sequence length $L=tn$, $m\ll n$, and the hidden dimension of local/global layers can be different.}
    \label{tab:complexity}
    \begin{tabular}{c|l|l}
    \toprule
         & Transformer (enc./dec. only) &  \modelname \\
    \midrule
       \multirow{2}{*}{Operating tensor(s)}  & \multirow{2}{*}{Data tokens: $\mathbb{R}^{L\times d}$} & Data tokens: $\mathbb{R}^{t\times n\times d}$ \\
       &&Latent tokens: $\mathbb{R}^{t\times m\times d}$\\ \hline
       \multirow{4}{*}{Attention layer complexity} & \multirow{4}{*}{$O(L^2d)$} &  $O(L^{\frac{4}{3}}d)$ (optimally)\\
       &&~- Local layer: $tn^2d$\\
       &&~- Global layer: $(tm)^2d$ \\
       &&~- Cross attention: $tnmd$ \\\hline
       \multirow{3}{*}{FFN layer complexity} & \multirow{3}{*}{$O(Ld^2)$} & $O(Ld^2)$\\
       &&~- Local layer: $tnd^2$\\ 
       &&~- Global layer: $tmd^2$ \\
    \bottomrule
    \end{tabular}
\end{table}

\begin{figure}[b]
    \centering
     \begin{subfigure}[b]{0.468\textwidth}
         \centering
         \includegraphics[width=\textwidth]{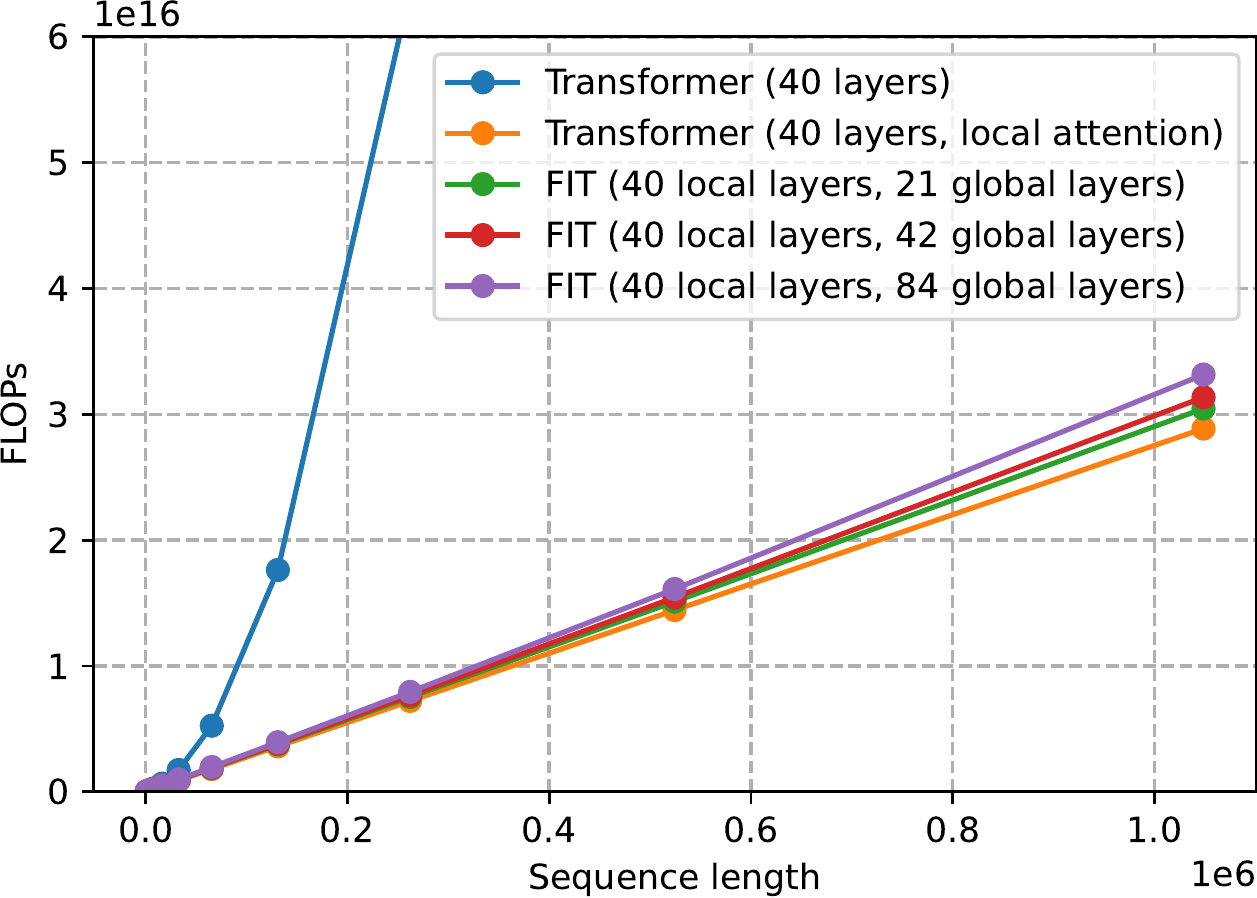}
         \caption{FLOPs scaling with sequence length.}
         \label{fig:case_saving1}
     \end{subfigure}
     \begin{subfigure}[b]{0.48\textwidth}
         \centering
         \includegraphics[width=\textwidth]{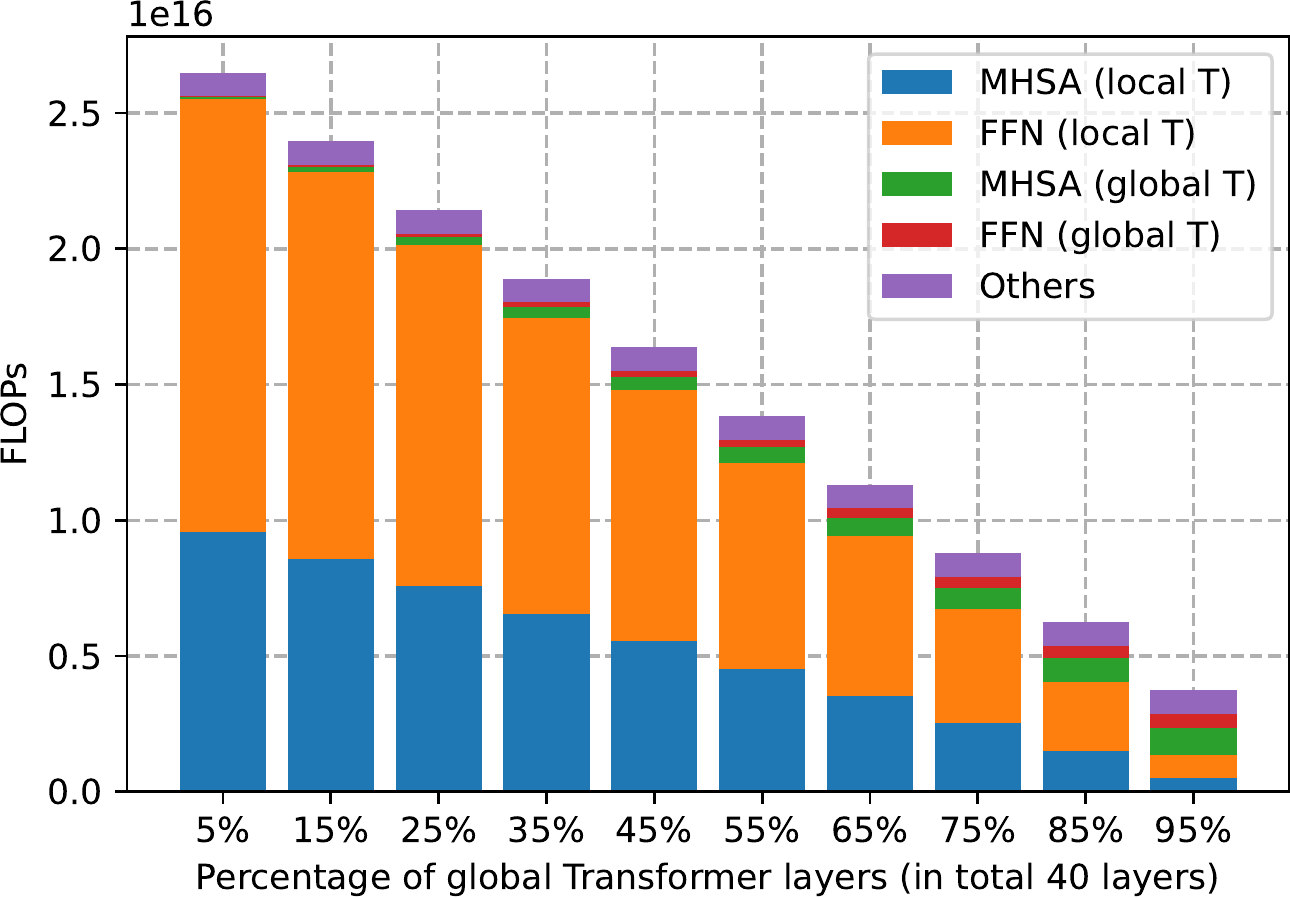}
         \caption{FLOPs scaling with global transformer layers.}
\label{fig:case_saving2}
     \end{subfigure}
    \caption{FLOPs analysis based on a $\sim$13B transformer model (40 layers with hidden dim of 5120), and for \modelname, the global latent tokens is set to 3\% of data tokens. (a) \modelname shares similar FLOPs as Transformers with \textit{only} local window attention, despite having extra global transformer layers (with even more parameters than local layers). (b) By relying more on the global transformer layers, we can further reduce the FLOPs, facilitated by the adaptive and compressive nature of the latent tokens. }
    \label{fig:case_saving}
\end{figure}

Figure~\ref{fig:case_saving} presents a case study focusing on the FLOPs (floating-point operations) analysis of a decoder-only transformer model with approximately 13 billion parameters. This model consists of 40 layers with a hidden dimension of 5120 (for both data and latent tokens). 
The group/window size is set to a fixed value of 2048 for all sequence lengths, resulting in a theoretical attention complexity of $O(L^2)$. However, with 64 latent tokens per group, the global attention operates on approximately 3\% of the data tokens.

In Figure~\ref{fig:case_saving1}, we observe that replacing the full attention of standard transformers with window/group attention can significantly reduce the FLOPs for long sequences. However, in this scenario, there is no global interaction among the groups. With \modelname, we incorporate global transformer layers to enable interactions across groups. Notably, thanks to the reduced set of latent tokens, the additional FLOPs required by the global transformer layers are relatively minimal, even for sequence lengths of 1 million tokens. Furthermore, Figure~\ref{fig:case_saving2} demonstrates that the computation in the local layers is much more computationally expensive compared to the global layers.
Consequently, by offloading the computation from the local layers to the global layers, further reductions in FLOPs can be achieved.
Similar analysis of a smaller 350M model and a larger 175B model can be found in appendix~\ref{app:flops_analysis}. \section{Experiments}

We evaluate the proposed architecture on three tasks: 1) high-resolution image understand via Pix2Seq object detection on object365~\cite{chen2021pix2seq,shao2019objects365}, 2) high-resolution image generation via pixel-based denoising diffusion models on Imagenet with 512$\times$512 or 1024$\times$1024 resolutions~\cite{ho2020denoising,jabri2022scalable,chen2023importance}, and 3) pixel-based autoregressive image generation on Imagenet-64$\times$64~\cite{van2016pixel}.

\subsection{High-resolution image understanding using Pix2Seq}

Pix2Seq~\cite{chen2021pix2seq,chen2022unified} is an approach that addresses various vision tasks, including object detection, segmentation, and keypoint detection, by employing an image-conditional language modeling framework. It utilizes an image encoder to extract meaningful visual features and a language decoder that generates object descriptions, such as bounding box coordinates and class labels. Here we evaluate \modelname alongside a commonly used vision transformer (ViT) encoder~\cite{dosovitskiy2020image}. Since Pix2Seq has a lower inductive bias, it benefits from pretraining on larger datasets like objects365~\cite{shao2019objects365}. Therefore, we follow similar settings as in~\cite{chen2021pix2seq} and assess the performance of different encoders using pretraining negative log-likelihood (Nll).

We partition the images into 16 sub-images, treating each sub-image as a group, and assign 32 latent tokens to each group. For comparison with ViT, we maintain a similar architecture but include a few additional global layers. As a result, the standard ViT layers now correspond to local layers operating independently within each group. Specifically, for FIT-B, we have L(4)G(2)L(4)G(2)L(4) layers, and for FIT-L, we have L(6)G(2)L(6)G(2)L(6)G(2)L(6) layers, where L/G represents local and global layers, respectively. Our experiments primarily focus on $640\times 640$ images to align with the settings in~\cite{chen2021pix2seq}, considering that ViT becomes computationally expensive at higher resolutions. However, we also provide a comparison of training efficiency for higher resolutions.

\begin{table}[h]
    \small
    \centering
    \caption{\label{tab:exp_detection}Comparison of visual encoders for Pix2Seq object detection on Objects365.}
    \subcaption{\label{tab:exp_detection1}On image resolution of 640$\times$640 (with patch size of 16$\times$16, there are 1600 data tokens in total).}
    \begin{tabular}{c|cccc}
    \toprule
    Image Encoder & Nll $\downarrow$ & Params (M) & FLOPs (G) & Steps per sec. $\uparrow$\\
    \midrule
    ViT-B & 2.49	& {123.2}	& 368 & 2.4  (1.0$\times$)\\
    FIT-B & {2.47}	& 161.4 & {332} & {3.4}  (1.4$\times$)\\ \midrule
    ViT-L & 2.37	& {342.7}	& 1223 & 0.9 (1.0$\times$)\\
    FIT-L & {2.35}	& 371.4	& {1036} & {1.6}  (1.8$\times$)\\
    \bottomrule
    \end{tabular}
    \subcaption{\label{tab:exp_detection2}Scaling up the image resolution. Using a single example per TPUv3 core.}
    \begin{tabular}{c|c|cccc}
    \toprule
    Image resolution&  & 640$\times$640 & 1024$\times$1024 & 1536$\times$1536 & 2048$\times$2048 \\ 
    \midrule
    Number of patches &  & 1600 & 4096 & 9216 & 16384 \\ 
    \midrule
    \multirow{2}{*}{FLOPs (G)}& ViT-B & 368 & 1326 & 4750 & 12840\\
    & FIT-B & 332 & 815 & 1900 & 3627\\
    \midrule
    \multirow{2}{*}{Steps per sec.}& ViT-B & 8.2 (1.0$\times$) & 2.5 (1.0$\times$) & 0.6 (1.0$\times$) & OOM  (n/a)\\
     & FIT-B & 9.2 (1.1$\times$) & 5.8 (2.3$\times$) & 2.8 (4.7$\times$) & 1.4 (n/a)\\
     \bottomrule
    \end{tabular}
\end{table}

The results are summarized in Table~\ref{tab:exp_detection}, and we observe that \modelname, by incorporating a few additional global layers, not only increases the steps per second during training but also achieves lowered loss. Furthermore, the speedup become more profound when we scale up the image resolutions. Notably, we are able to train a $>$300M FIT model on TPUv3, without the need for special optimization or model parallelism, to handle 6400$\times$6400 resolution images, which contain $6400\times 6400\times 3\times 8$ bits ($\sim$1GB) of raw input data. 
Since the image is tokenized into $16\times 16$ patches, the resulting input consists of 160K tokens.

\subsection{Pixel-based end-to-end diffusion modeling}

RIN~\cite{jabri2022scalable,chen2023importance}, a recent advancement in architectural design and modeling for denoising diffusion models~\cite{ho2020denoising}, has demonstrated the ability to train directly on high-resolution images up to 1024$\times$1024. As mentioned earlier, RIN can be viewed as a specific instance of the basic \modelname architecture with a single group of tokens and without self-attention in the local layers. Since RIN is optimized for the diffusion models, we directly compare \modelname to RIN by incorporating additional groups. For the evaluation of diffusion model training, we utilize the mean square error (MSE) between predictions and targets as the performance metric for different architectural choices.

\begin{table}[h]
    \small
    \centering
    \caption{\label{tab:exp_diffusion}Comparison of denoising diffusion training between a \modelname (removing self-attention on local layers) and RIN~\cite{jabri2022scalable} (equivalent to group=1). We see that by spiting images into sub-images and treat each sub-image as a group for cross-attention between data and corresponding latent tokens, the MSE is reduced while the training efficiency is improved.}
    \begin{tabular}{c|ccc|ccc}
    \toprule
    & \multicolumn{3}{c|}{512$\times$512 resolution} & \multicolumn{3}{c}{1024$\times$1024 resolution}\\
    \midrule
    Groups & MSE ($\times 1e^{-3}$) & FLOPs (G) & Steps per sec. & MSE ($\times 1e^{-3}$) & FLOPs (G) & Steps per sec. \\\midrule
    1 (RIN) & 2.79 & 344 & 1.5  (1.0$\times$) & 0.80 & 939 & 0.9 (1.0$\times$) \\
    4 & 2.74 & 324 & 1.9  (1.3$\times$)& 0.77 & 858 & 1.3 (1.4$\times$) \\
    16 & {2.73} & {318} & {2.2}  (1.5$\times$) & {0.75} & 838 & 1.5  (1.7$\times$)\\
    64 & {2.73} & {318} & 1.9  (1.3$\times$) & 0.76 & {833} & {1.6} (1.8$\times$) \\
    \bottomrule
    \end{tabular}
\end{table}

The summarized results can be found in Table~\ref{tab:exp_diffusion}, and it is evident that by incorporating additional groups and transforming RIN into \modelname, we observe a noticeable decrease in mean square error (MSE) and a significant improvement in training speed measured in steps per second (on TPUv3).

\subsection{Pixel-based image autoregressive modeling}

Modeling pixels directly as discrete tokens in an autoregressive manner~\cite{van2016pixel,chen2020generative} presents challenges due to the long sequence length (e.g., 64x64 images result in 12,288 data tokens) and the need to capture both local and global dependencies. In our approach, we group pixels locally into 8$\times$8 patches, resulting in 192 data tokens per group, and utilize 32 latent tokens per group. We use 512d for data token and 768d for latent token, with layer configuration of L(8)G(2)L(8)G(2)L(8)G(2)L(8). The summarized results in Table~\ref{tab:exp_ar} demonstrate near state-of-the-art performance, considering the model's size.

\begin{table}[h]
    \small
    \centering
    \caption{\label{tab:exp_ar}Comparison among different pixel-level autoregressive models on ImageNet-64$\times$64.}
    \begin{tabular}{c|ccc}
    \toprule
    Method & Params (M) & Bits per dim $\downarrow$ \\
    \midrule
    PixelCNN~\cite{van2016pixel} & - & 3.57 \\
    PixelSNAIL~\cite{chen2018pixelsnail} & - & 3.52 \\
    SPN~\cite{menick2018generating} & - & 3.52 \\
    Reformer~\cite{kitaev2020reformer} & - & 3.65 \\
    Image transformer~\cite{parmar2018image} & - & 3.48 \\
    Sparse transformer~\cite{child2019generating} & 152 & 3.44 \\
    Routing transformer~\cite{roy2021efficient}	& >200 & 3.43 \\
    Combiner~\cite{ren2021combiner} & 249 & 3.42\\
    Perceiver AR~\cite{hawthorne2022general} & 770 & 3.40 \\
    \midrule
    FIT & 153 & 3.42 \\
    \bottomrule
    \end{tabular}
\end{table}

\subsection{Ablation study}

Table~\ref{tab:exp_ab_latents} showcases the effectiveness of utilizing a larger number of latents. Interestingly, we observe that increasing the number of latents does not necessarily result in a significant increase in parameters or training time (steps per second), particularly when the local layers contribute more to the computation.

\begin{table}[h]
    \small
    \centering
    \caption{Effects of number of latents. Increasing number of latents has positive effect on negative log-likelihood (nll) and bit per dim (bpd), while having minor effects on parameters and run time. Note the latent tokens are still much smaller than data tokens (percentage shown in the parentheses).}
    \label{tab:exp_ab_latents}
    \begin{tabular}{cccc|cccc}
    \toprule
    \multicolumn{4}{c|}{Object365 (Pix2Seq)} & \multicolumn{4}{c}{Imagenet-64$\times$64 (Autoregressive)}\\
    \midrule
    Num. latents & Nll & Params (M) & Steps per sec. & Num. latents & Bpd & Params (M) & Steps per sec. \\\midrule
    16$\times$4 (4\%) & 2.66 & 154 & 2.03 & 64$\times$16 (8\%) & 3.46 & 144 & 1.2 \\
    16$\times$16 (16\%) & 2.63 & 154 & 1.96 & 64$\times$32 (17\%) & 3.45 & 145 & 1.0 \\
    \bottomrule
    \end{tabular}
\end{table}

Table~\ref{tab:exp_ab_interleave} investigates various layer interleaving patterns while maintaining a constant number of local and global layers (except for the case where only local layers are present, which corresponds to the original local and global layers) and we use the same hidden dimension for both types of layers. Notably, we observe that interleaving local and global layers is crucial for achieving optimal results, while maintaining roughly the same training efficiency (measured in steps per second on TPUv3).

\begin{table}[h]
    \small
    \centering
    \caption{\label{tab:exp_ab_interleave} Comparison across various layer interleave patterns. We use the same hidden dimension for both local and global layers and keep the total number of local/global layers constant (except for the case labeled as L, which only consists of local layers). Interleaving the local and global layers yields improved performance with negligible impact on training cost (steps per second).
    }
    \begin{tabular}{c|cccc}
    \toprule
    Task & Interleave pattern & Nll or Bpd $\downarrow$ & Params (M) & Steps per sec.\\
    \midrule
    \multirow{4}{*}{\shortstack{Object365\\ (Pix2Seq)}} & L	& 2.66 & 138 & 2.1\\
    & L$\rightarrow$G$\rightarrow$L	& 2.52 & 157 & 2.0\\
    & L$\rightarrow$G$\rightarrow$L$\rightarrow$G$\rightarrow$L & 2.51 & 161 & 2.0\\
    & L$\rightarrow$G$\rightarrow$L$\rightarrow$G$\rightarrow$L$\rightarrow$G$\rightarrow$L & 2.50 & 173 & 1.9\\
    \midrule
    \multirow{4}{*}{\shortstack{Imagenet-64$\times$64\\ (Autoregressive)}}  & L	& 3.80 & 45 & 1.9\\
    & L$\rightarrow$G$\rightarrow$L & 3.51 & 48 & 2.4\\
    & L$\rightarrow$G$\rightarrow$L$\rightarrow$G$\rightarrow$L & 3.50 & 49 & 2.4\\
    & L$\rightarrow$G$\rightarrow$L$\rightarrow$G$\rightarrow$L$\rightarrow$G$\rightarrow$L	& 3.49 & 52 & 2.3\\
    \bottomrule
    \end{tabular}
\end{table} \section{Conclusion}

We introduced FIT, or FitTransformer. On one hand, \modelname can be viewed as connecting local transformers that operate independently on different groups or segments of data tokens with global transformers that provide contextualized feedback. On the other hand, \modelname can also be seen as enabling global transformers with learned tokenization through a set of latent tokens that selectively attend to data tokens, resulting in more adaptive computation.
As a result, \modelname has the capability to process raw input data of nearly 1GB in size during training, which, if proven effective, we believe opens up new opportunities and potential applications in the future.
It is important to note that our empirical study is preliminary, and additional evaluation is necessary to ascertain the applicability of this architecture as a transformer surrogate for sequences of varying sizes, or how it should be further refined.
Additionally, while we have primarily applied \modelname to image understanding and generation, it is versatile and can also be adapted to other domains, such as video and text.

\section*{Acknowledgements}
We specially thank Geoffrey Hinton, Ruoxi Wang, David Fleet, Mahesh Sathiamoorthy for helpful discussion and feedback on the draft.

{\small
\bibliography{content/ref}
\bibliographystyle{plainnat}}

\clearpage
\newpage
\appendix

\section{Extra pseudo-code}

\begin{figure}[ht]
  \centering
\begin{minipage}[t]{0.9\textwidth}
\begin{algorithm}[H]
\small
\caption{Pseudo-code for utility functions used.}
\label{alg:utils}
\definecolor{codeblue}{rgb}{0.25,0.5,0.5}
\definecolor{orange}{rgb}{0.96, 0.81, 0.375}
\definecolor{codekw}{rgb}{0.85, 0.18, 0.50}
\lstset{
  backgroundcolor=\color{white},
  basicstyle=\fontsize{8pt}{8pt}\ttfamily\selectfont,
  columns=fullflexible,
  breaklines=true,
  captionpos=b,
  commentstyle=\fontsize{8pt}{8pt}\color{codeblue},
  keywordstyle=\fontsize{8pt}{8pt}\color{codekw},
  escapechar={|}, 
}
\begin{lstlisting}[language=python]
def local/global_transformers(x, mask=None, use_attn=true, use_ffn=true, layers=K):
  """x in (b, n, c) and multi-head attention is across n tokens."""
  for k in range(K):
    if use_attn:
      x += multihead_attention[k](q=x, k=x, v=x, attn_mask=mask)
    if use_ffn:
      x += ffn[k](x)
  return x
  
def x2l/l2x_cross_attn(x, y):
  """x in (b, n, c), y in (b, m, d), attention is across n|$\times$|m tokens."""
  x += multihead_attention(q=x, k=y, v=y)
  return x
  
def shift_latents(latents):
  """Shift latents (b, t, m, d) by 1 group to the right, and pad in the front."""
  latents_leading, latents_last = latents[:, :-1], latents[:, -1:]
  latents = concat([zeros_like(latents_last), latents_leading], axis=1)
  return rearrange(latents, `b t m d -> (b t) m d`), latents_last

def shift_back_latents(latents, latents_last):
  """Remove the padding group and restore the last group."""
  latents = rearrange(latents, `(b t) m d -> b t m d`)
  return concat([latents[:, 1:], latents_last], axis=1)
\end{lstlisting}
\end{algorithm}
\end{minipage}
\end{figure}

\section{Computational complexity analysis for attention layers in \modelname}
\label{app:attn_complexity}

Here we delve into a more detailed complexity analysis of \modelname's attention layers. We will exclude the cross attention between data and latent tokens, as its complexity is linear with respect to the sequence length.
Consider a sequence of length $L$, which we divide into $t$ groups. Each group contains $n$ data tokens, resulting in $L = tn$. We assume a constant number of latent tokens per group. The attention complexity for each local transformer layer is $O(tn^2)$, while the attention complexity in a global transformer layer is $O(t^2)$. Therefore, the overall complexity of a single local layer and a single global layer, which are treated as an approximation to the full attention in a standard transformer layer, can be expressed as $c_1 tn^2 + c_2 t^2 = c_1 Ln + c_2 (L/n)^2$.

If we maintain a constant value for $n$, such as 1024, regardless $L$, the attention complexity remains $O(L^2)$. However, due to a reduced constant factor, the computational efficiency can still be orders of magnitude faster than the standard full attention.

Alternatively, we can allow $n$ to vary as a function of $L$. In this case, the optimal choice is $n = (\frac{2c_2}{c_1} L)^\frac{1}{3}$. With this selection, the overall attention complexity becomes $O(n^\frac{4}{3})$, which represents a significant reduction compared to the $O(L^2)$ of standard full attention.

\section{FLOPs analysis based on GPT-3 Medium (350M) and GPT-3 (175B)}
\label{app:flops_analysis}

\begin{figure}[t]
    \centering
     \begin{subfigure}[b]{0.48\textwidth}
         \centering
         \includegraphics[width=\textwidth]{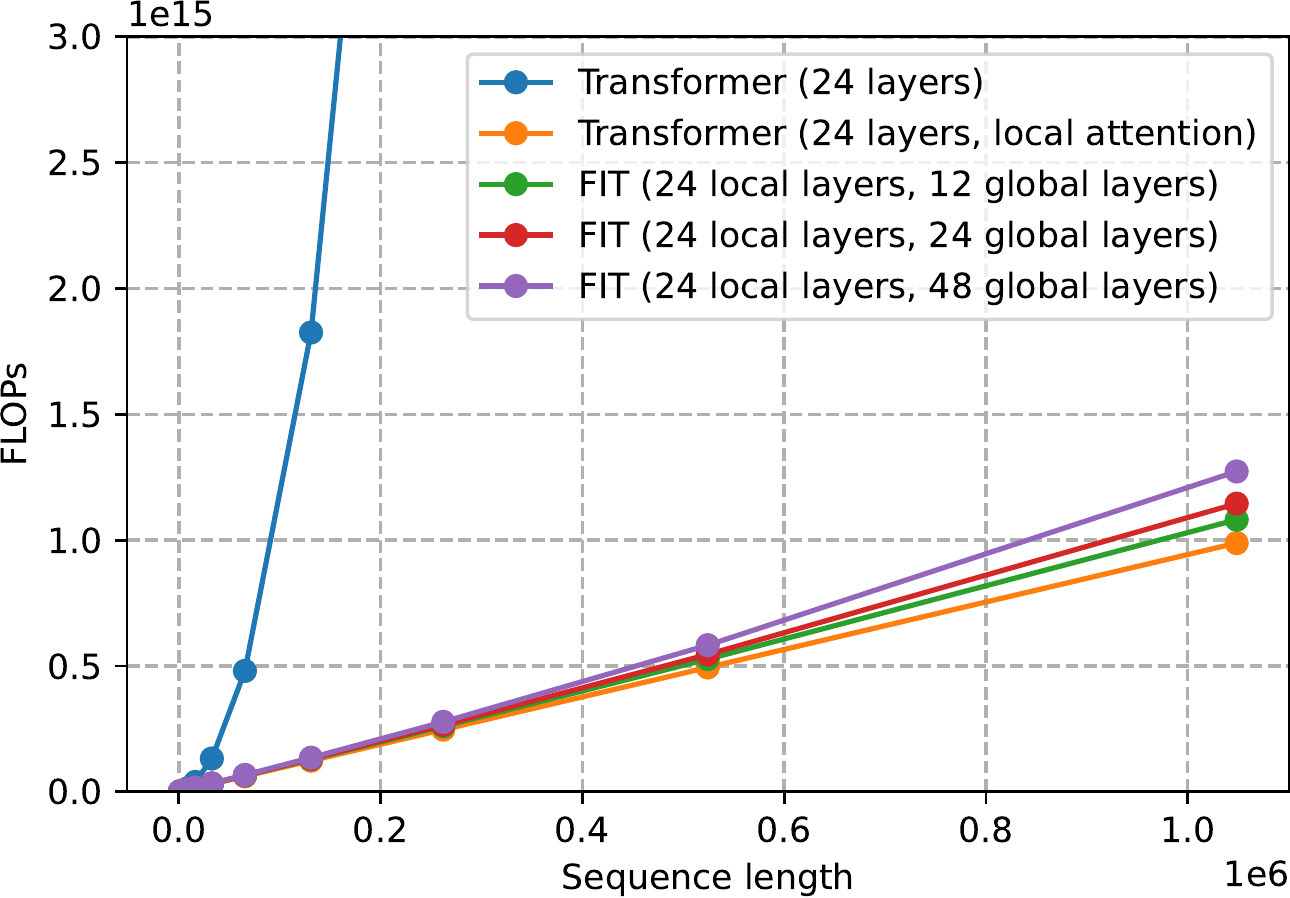}
         \caption{GPT-3 Medium (350M).}
         \label{fig:case_saving1}
     \end{subfigure}
     \begin{subfigure}[b]{0.471\textwidth}
         \centering
         \includegraphics[width=\textwidth]{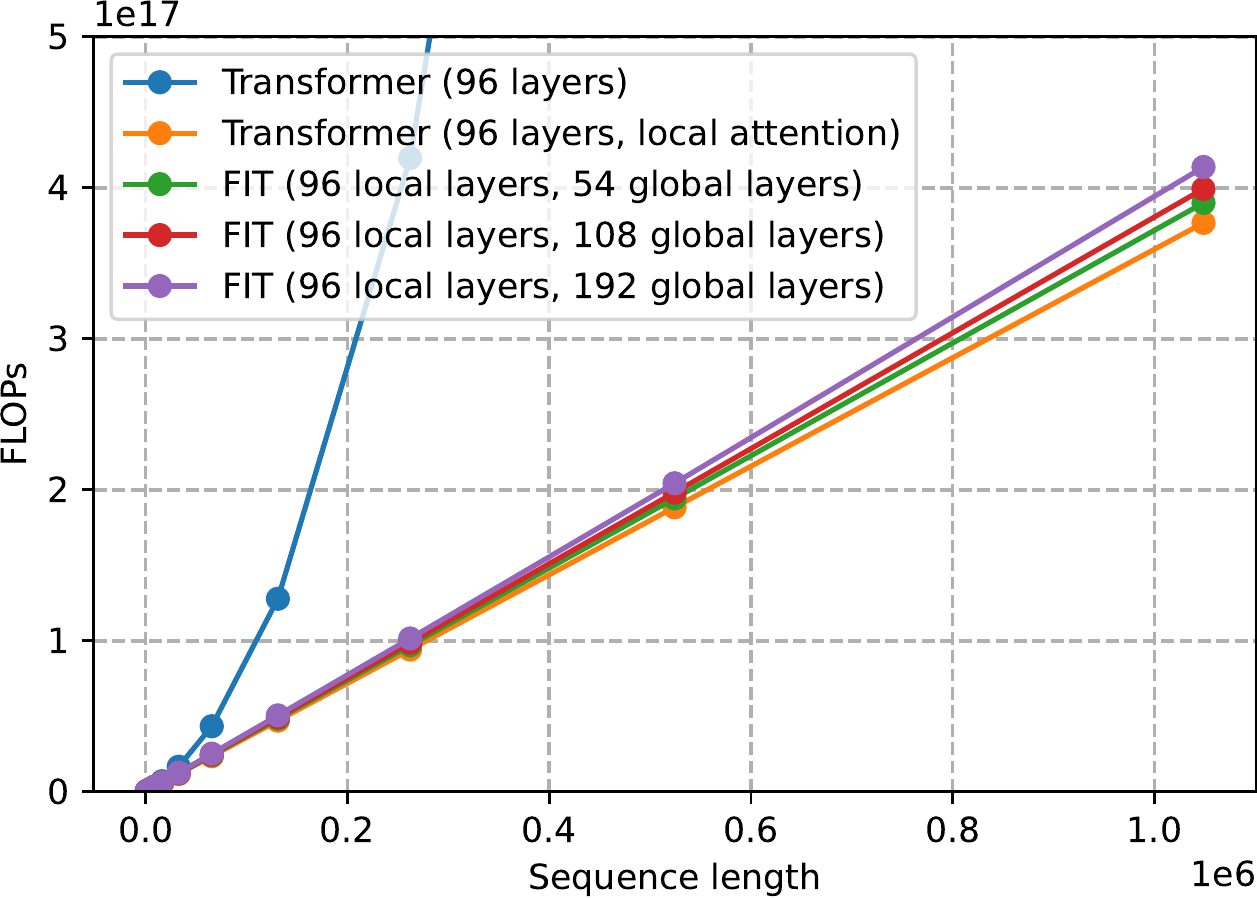}
         \caption{GPT-3 (175B).}
\label{fig:case_saving2}
     \end{subfigure}
    \caption{FLOPs scaling with sequence length, based on GPT-3 Medium (350M, 1024d, 24 layers) and GPT-3 (175B, 12288d, 96 layers). Similar to that of 13B model in Figure~\ref{fig:case_saving}, \modelname shares similar FLOPs as Transformers with \textit{only} local window attention, despite having extra global transformer layers (with even more parameters than local layers).
    }
    \label{fig:case_saving_more}
\end{figure}

In Figure~\ref{fig:case_saving_more}, we analyze the floating-point operations (FLOPs) of the GPT-3 Medium (350M), GPT-3 (175B) models, and FIT based on them. We maintain a group/window size of 2048 for all sequence lengths and use 64 latent tokens per group. The results remain consistent with the previous findings, indicating that the FIT model (with additional global layers, even double the parameter count) closely aligns with the GPT models using window attention only (in terms of FLOPs).

\section{Overlapped grouping}
\label{app:overlap_group}

\begin{figure}[t]
    \centering
    \includegraphics[width=0.9\textwidth]{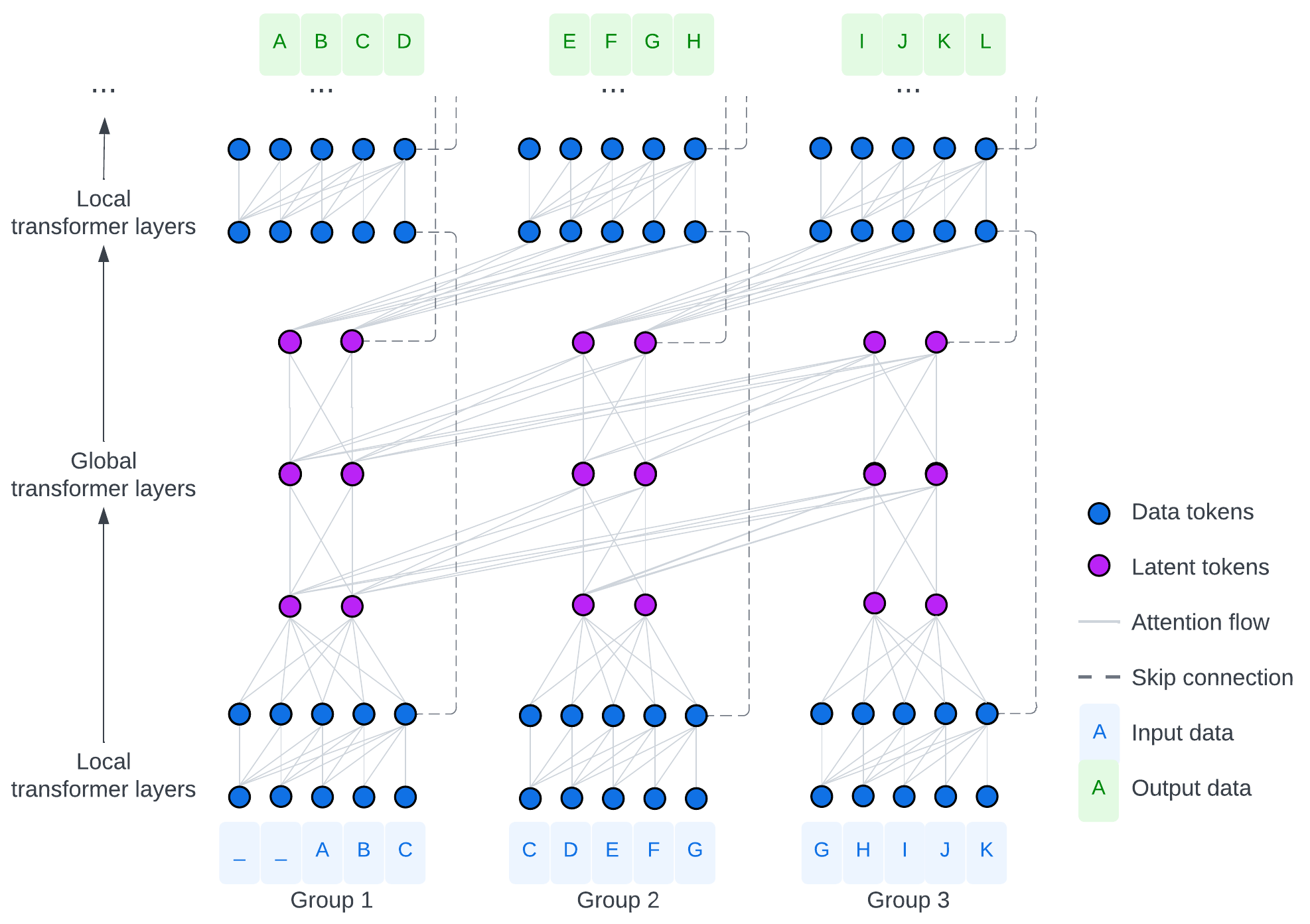}
    \caption{\label{fig:fit_ar_overlap}Illustration of the \modelnamear with overlapping between adjacent groups / segments. ``C'' and ``G'' in group 2 and 3 are ``prefix tokens'', whose next token predictions are discarded.
    }
\end{figure}

A concern that arises from using non-overlapping token grouping is the potential separation of adjacent and highly dependent tokens into different groups. To tackle this issue, one can utilize overlapped grouping. Take autoregressive modeling as an example, one can optionally include a few ``prefix tokens'' from the previous group, as shown in Figure~\ref{fig:fit_ar_overlap}. These ``prefix tokens'' are solely used for conditioning purposes, and the corresponding next-tokens generated by them will be disregarded.

\section{More discussions on closely related work}
\label{app:related}

We highlight some differences of \modelname w.r.t. the closely related architectures in the literature.

\textbf{BigBird / ETC / Longformer}~\cite{zaheer2020big,ainslie2020etc,beltagy2020longformer}. 
These methods also leverage latent/memory tokens and incorporate both local attention among data tokens and global attention between memory and data tokens. However, instead of replacing a standard full attention layer with factorized attention layers in these methods, we separate the two types of attentions into distinct transformer layers that we interleave. This offers greater flexibility in specifying the computation distribution for each type and minimizes additional operations such as tensor reshaping. Additionally, for cross attention between latent and data tokens, we limit it to within each group rather than performing it globally, which reduces the cost of cross attention and allows for using a relatively larger number of latents. Unlike these prior architectures, we adopt a simple grouping of tokens and utilize standard self-attention within each group, avoiding the less efficient sliding window local attention. Lastly, it is important to note that these prior architectures do not support autoregressive generation, which is a significant application in language modeling.

\textbf{Perceiver (IO) / Perceiver AR}~\cite{jaegle2021perceiver,jaegle2021perceiverio,hawthorne2022general}. 
Both Perceiver (IO) and Perceiver AR utilize latent tokens and cross attention to facilitate communication between latent and data tokens. However, they operate on a single group consisting of all data tokens without significant processing (e.g. self-attention) for the data tokens. We have observed the significance of local processing, and the absence of such processing can present challenges in effectively routing information through cross attention. Additionally, in Perceiver (IO) and Perceiver AR, the data tokens are either not updated during the forward network after initial formation or updated only once at the end of the forward network. In contrast, \modelname utilizes an interleaved mechanism to iteratively update the data tokens throughout the forward pass. This iterative update allows for ongoing refinement of the data tokens from global context and enables richer information flow across longer range.

\textbf{Hierarchical Perceiver (HiP)}~\cite{carreira2022hierarchical}. 
HiP is based on Perceiver IO and incorporates groups and multiple levels of latents, whereas \modelname maintains simplicity by utilizing a single level of latents. Similar to Perceiver IO, HiP forgoes local processing (e.g., self-attention) for data tokens and only updates the data tokens once at the end of the forward pass, which faces similar potential drawbacks as in Perceiver IO. Additionally, like Perceiver IO, HiP does not support autoregressive modeling.

\textbf{RIN}~\cite{jabri2022scalable}. 
As mentioned earlier, RIN can be considered a special case of basic \modelname architecture. RIN operates with a single group and lacks local self-attention, which may limit its capacity as all communication between data tokens relies on a small set of latents. In contrast, \modelname introduces multiple groups, enabling support for autoregressive modeling and reducing the computational cost of cross attention between latent and data tokens. These advancements in \modelname make it a crucial improvement over RIN.

\textbf{MEGABYTE}~\cite{yu2023megabyte} (concurrent work). 
Both approaches share some similar terminologies such as global and local transformers, but they diverge in several key aspects. 

MEGABYTE utilizes a single global transformer model followed by a local transformer to directly predict the data output. In contrast, \modelname incorporates interleaved local and global transformer layers in multiple architectural blocks, enabling modulation between local and global processing, which we have found to improve performance.

The global transformer layers in MEGABYTE operate directly on patch embedding tokens. Consequently, when MEGABYTE is used as an image encoder, it resembles ViT~\cite{dosovitskiy2020image} followed by an extra local transformer. In contrast, the local transformer layers in \modelname can also operate on patch embedding tokens, similar to the global transformer in MEGABYTE. Moreover, the global layers in \modelname operate on a set of introduced latent tokens that can adaptively aggregate information from patch embedding tokens. This distinction allows \modelname to handle even larger input data, such as 6400$\times$6400 images, compared to MEGABYTE's 640$\times$640. 

Despite these differences, the idea presented in MEGABYTE can potentially be integrated into \modelnamear by adding an additional local autoregressive transformer. This local transformer, similar to that in MEGABYTE, would be responsible for decoding the output data from each data token. 
\end{document}